\documentclass[conference]{IEEEtran}
\IEEEoverridecommandlockouts
\usepackage{booktabs}
\newcommand{\pmtext}[1]{\text{$\pm$\scriptsize #1}}
\usepackage{cite}
\usepackage{amsmath,amssymb,amsfonts}
\usepackage{algorithmic}
\usepackage{graphicx}
\usepackage{textcomp}
\usepackage{xcolor}
\def\BibTeX{{\rm B\kern-.05em{\sc i\kern-.025em b}\kern-.08em
    T\kern-.1667em\lower.7ex\hbox{E}\kern-.125emX}}
\begin{document}

\title{Mamba2MIL: State Space Duality Based Multiple Instance Learning for Computational Pathology\\
\thanks{*Corresponding author}
}

\author{
    \IEEEauthorblockN{1\textsuperscript{st} Yuqi Zhang}
    \IEEEauthorblockA{
        \textit{School of Computer Science {$\&$}  Engineering} \\
        \textit{State Key Laboratory of Virtual Reality}\\
        \textit{Technology and Systems}\\
        \textit{Beihang University,}
        Beijing, China\\
        yqzhang@buaa.edu.cn}
    \\
    \IEEEauthorblockN{3\textsuperscript{rd} Jiakai Wang}
    \IEEEauthorblockA{
        \textit{Zhongguancun Laboratory}\\
        Beijing, China \\
        wangjk@zgclab.edu.cn}
    \\
    \IEEEauthorblockN{5\textsuperscript{th} Taiying Peng}
    \IEEEauthorblockA{
        \textit{School of Computer Science {$\&$}  Engineering} \\
        \textit{Beihang University,}
        Beijing, China\\
        taiyi@buaa.edu.cn}
    \and
    \IEEEauthorblockN{2\textsuperscript{nd} Xiaoqian Zhang}
    \IEEEauthorblockA{
        \textit{School of Biological Science and Medical Engineering}\\
        \textit{Beihang University,}
        Beijing, China\\
        shawnozhang@outlook.com}
    \\
    \\
    \IEEEauthorblockN{4\textsuperscript{th} Yuancheng Yang}
    \IEEEauthorblockA{
        \textit{School of Computer Science {$\&$}  Engineering} \\
        \textit{State Key Laboratory of Virtual Reality}
        \textit{Technology and Systems}\\
        \textit{Beihang University,}
        Beijing, China\\
        ycyoung@buaa.edu.cn}
    
    \\
    \IEEEauthorblockN{6\textsuperscript{th} Chao Tong*}
    \IEEEauthorblockA{\textit{School of Computer Science {$\&$}  Engineering} \\
        \textit{State Key Laboratory of Virtual Reality}
        \textit{Technology and Systems}\\
        \textit{Beihang University,}
        Beijing, China\\
        tongchao@buaa.edu.cn}
}

\maketitle

\begin{abstract}
    Computational pathology (CPath) has significantly advanced the clinical practice of pathology. Despite the progress made, Multiple Instance Learning (MIL), a promising paradigm within CPath, continues to face challenges, particularly related to incomplete information utilization. Existing frameworks, such as those based on Convolutional Neural Networks (CNNs), attention, and selective scan space state sequential model (SSM), lack sufficient flexibility and scalability in fusing diverse features, and cannot effectively fuse diverse features. Additionally, current approaches do not adequately exploit order-related and order-independent features, resulting in suboptimal utilization of sequence information. To address these limitations, we propose a novel MIL framework called Mamba2MIL. Our framework utilizes the state space duality model (SSD) to model long sequences of patches of whole slide images (WSIs), which, combined with weighted feature selection, supports the fusion processing of more branching features and can be extended according to specific application needs. Moreover, we introduce a sequence transformation method tailored to varying WSI sizes, which enhances sequence-independent features while preserving local sequence information, thereby improving sequence information utilization. Extensive experiments demonstrate that Mamba2MIL surpasses state-of-the-art MIL methods. We conducted extensive experiments across multiple datasets, achieving improvements in nearly all performance metrics. Specifically, on the NSCLC dataset, Mamba2MIL achieves a binary tumor classification AUC of 0.9533 and an accuracy of 0.8794. On the BRACS dataset, it achieves a multiclass classification AUC of 0.7986 and an accuracy of 0.4981. The code is available at https://github.com/YuqiZhang-Buaa/Mamba2MIL.
   \end{abstract}


\section{Introduction}
Pathology plays a vital role in medicine, providing the scientific basis for diagnosing, treating, and prognosticating diseases by studying their nature, causes, development, and cellular effects. The clinical practice of pathology involves various tasks, including tumor detection, subtype classification, and staging\cite{chen2024towards, song2023artificial}. Given the urgency of these diagnoses and treatments, pathologists must adeptly solve diverse problems in a limited timeframe\cite{lipkova2022artificial}. However, the meticulous examination of individual cells or cell clusters under a microscope (or within gigapixel images) for malignancies can be exceedingly time-consuming and labor-intensive, imposing a significant burden on clinical practice\cite{jiang2023deep}.

\begin{figure}[!t]
    \centering
    \begin{minipage}[t]{1.0\linewidth}
    \centering
        \begin{tabular}{@{\extracolsep{\fill}}c@{}c@{}@{\extracolsep{\fill}}}
            \includegraphics[width=0.5\linewidth]{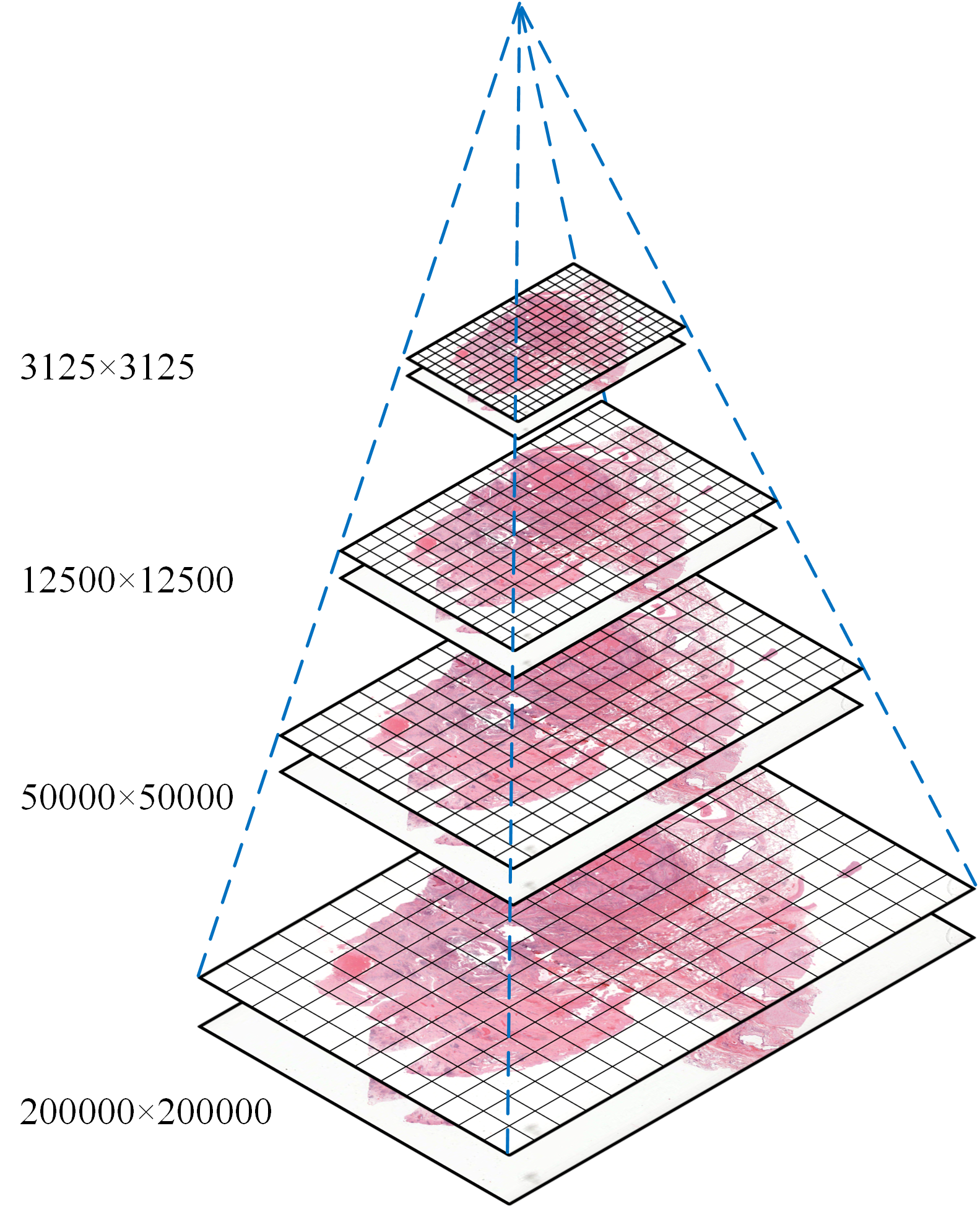} &
            \includegraphics[width=0.5\linewidth]{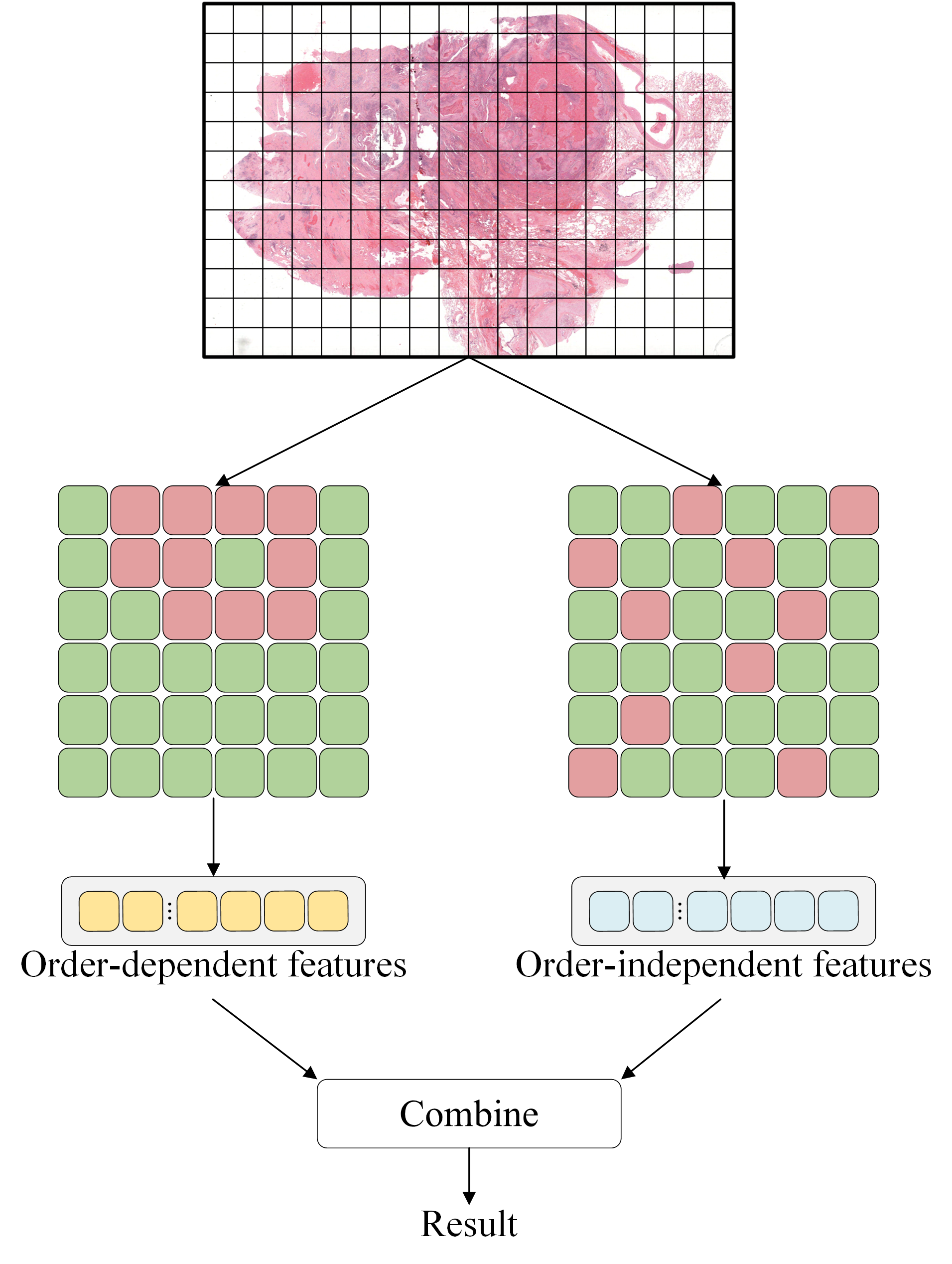}\\
            \footnotesize(a) & \footnotesize(b)\\
        \end{tabular}
    \end{minipage}
    \caption{The structure of WSI and our model pipeline. (a)An image pyramid structure of WSI. (b)Combine order-dependent and order-independent features.}
    \label{fig1}
 \end{figure}

The advent of computational pathology (CPath) offers a promising solution to this challenge\cite{kather2019deep, chen2022pan}. CPath leverages advanced computational techniques to analyze and interpret whole slide images (WSIs) in digital pathology, assisting pathologists in diagnosing and predicting diseases. The structure of WSIs is shown in Figure \ref{fig1}-a. Recently, the integration of artificial intelligence (AI) technologies has revolutionized CPath, with deep learning significantly enhancing the automation and accuracy of pathology image analysis\cite{jiang2023deep}. Furthermore, advancements in scanning systems, imaging technologies, and storage solutions have led to increased acquisition of WSIs in clinical settings, facilitating the growth of CPath.

Multiple Instance Learning (MIL)\cite{amores2013multiple} has emerged as an ideal framework for CPath, particularly in the analysis of WSIs. In this approach, each WSI (known as a bag) is divided into a large collection of patches (known as instances) which can number in the tens of thousands. The underlying assumption of MIL is that if at least one patch in the collection is categorized as positive, the entire WSI should be classified as positive. Conversely, if all patches are classified as negative, the WSI is deemed negative. The prevailing MIL paradigm involves transforming these patches into low-dimensional feature representations using pre-trained models on pathology images. These low-dimensional features are then aggregated into bag-level representations for further analysis. Despite its promise, current MIL approaches are limited by incomplete information utilization.  On one hand, they struggle to effectively fuse diverse features. Existing frameworks based on convolutional neural networks (CNN)\cite{kanavati2020weakly, cao2023e2efp}, attention\cite{li2021dt, lu2021data, li2021dual}, and selective scan space state sequential model (SSM)\cite{gu2023mamba, yang2024mambamil, fang2024mammil} often have fixed structures, which limits their flexibility and scalability in effectively fusing diverse features from high-resolution WSIs. On the other hand, sequence features are not effectively leveraged. The attention-based methods are well-aligned with the MIL assumption, which means, the sequence order of WSI patches should not affect the outcomes by principle. However, in practice, contextual information between patches plays a crucial role, and current methods do not fully exploit both order-dependent and order-independent features, as shown in Figure \ref{fig1}-b.

To address the challenges outlined above, we propose a novel MIL framework named Mamba2MIL. Our framework leverages state space duality model (SSD)\cite{dao2024transformers} for effective long-sequence modeling of WSI patches, combined with feature-weighted selection using hyperbolic tangent activation functions. This framework supports the fusion of a broader range of features and can be extended as needed according to specific usage requirements. Additionally, we introduce a streamlined sequence transformation method specifically designed for WSIs. This method is not only efficient and adaptable to different WSI sizes, but also enhances sequence-independent features while preserving local sequence information, thereby improving the overall utilization of sequence information. Our contributions can be summarized as follows:  
\begin{itemize}
    \item To the best of our knowledge, we are the first to propose an extensible SSM-based MIL framework that supports the fusion of a wider range of features.
    \item We introduce a sequence transformation method that further enhances the utilization of sequence information while maintaining efficiency across different WSI sizes.
    \item We conducted extensive experiments across multiple datasets, achieving improvements in nearly all performance metrics compared to state-of-the-art methods.
    \end{itemize}

 \begin{figure*}[!t]
    \centering
    \includegraphics[width=0.8\linewidth]{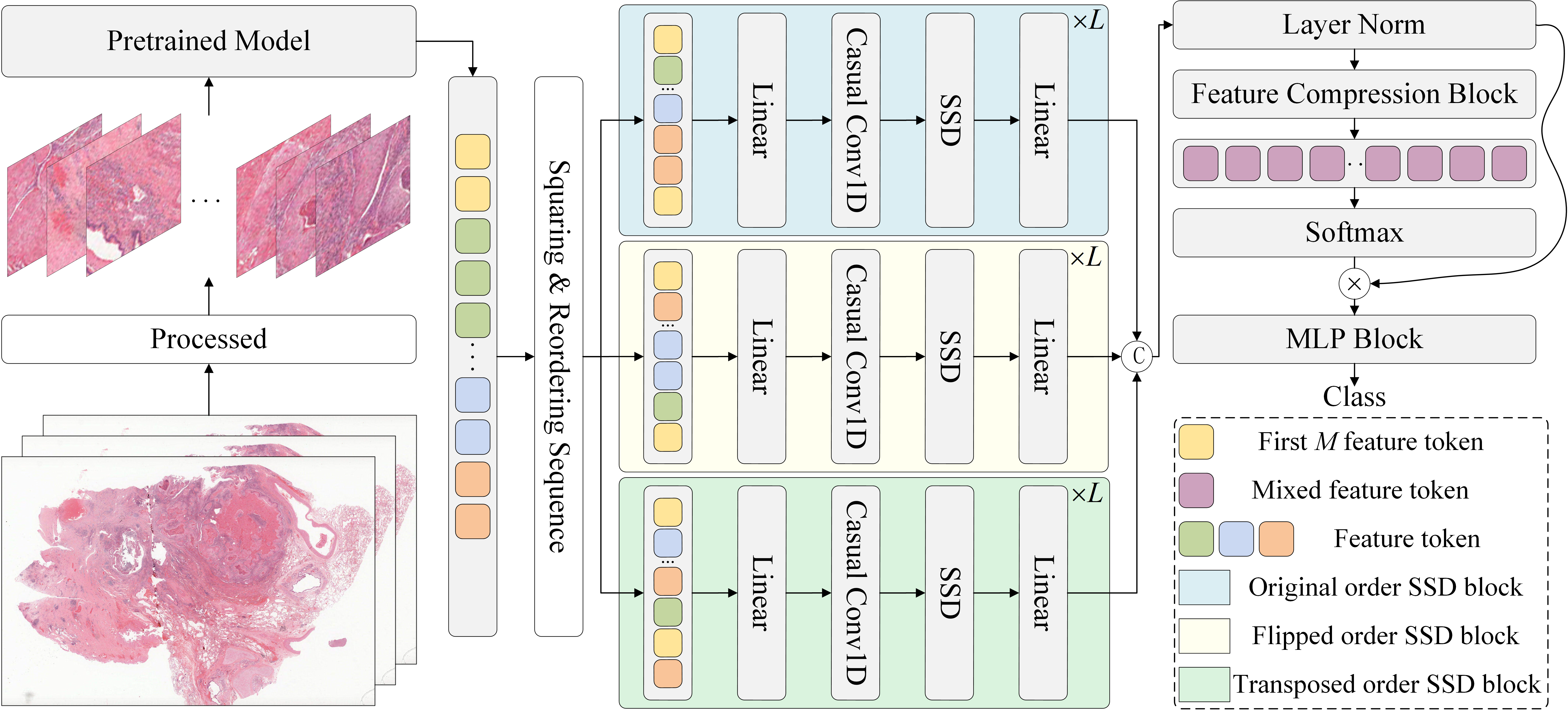}
    \caption{Overview of our model. Each WSI is first preprocessed to exclude the background and then divided into patches. These patches are embedded into feature vectors using a pretrained ResNet50 model. The resulting patch sequences undergo squaring and reordering operations to create differently ordered sequences. These three patch sequences are then processed through the SSD block, the feature selection block and an MLP block, to produce the final result.}
    \label{fig2}
  \end{figure*}

\section{Related Work}
\subsection{Multiple instance learning in CPath}
Clinical practice relies heavily on pathologists to manually annotate and analyze pathology images. This traditional approach is not only time-consuming and labor-intensive but also prone to errors such as missed diagnoses due to the large size and complexity of pathology images. To address these challenges, researchers have recently proposed weakly supervised learning methods based on MIL\cite{van2021deep, shmatko2022artificial}. MIL enables models to learn and predict pathology results from WSIs without the need for pixel-level annotations. MIL in CPath can be broadly categorized into two main approaches: instance-level algorithms\cite{lerousseau2020weakly, chikontwe2020multiple, kanavati2020weakly} and embedding-level algorithms. The instance-level algorithm trains the network with pseudo-labels assigned to each instance, where the pseudo-labels are based on bag-level labels. Subsequently, the top $k$ instances, ranked by cancer probability, are selected for aggregation. However, this method has limitations, such as a strong dependence on the number of WSIs and limited applicability for differentiating multiple tumor subtypes.
The second and more widely adopted algorithm is embedding-level algorithms\cite{li2023task, lu2023visual, song2024morphological}. These algorithms utilize a pretrained pathology image model to map each patch in the entire WSI into a fixed-length embedding space. The aggregated feature embeddings are then processed by an operator to obtain the final classification result. Among these, the CLAM\cite{lu2021data} model proposed by Lu et al. has set a standard for WSI processing, gaining widespread recognition and having a significant impact in the field. In this paper, we follow the same data processing flow as the CLAM model.

\subsection{Methods for combining order-dependent and order-independent features in MIL}
The above MIL methods based on the assumption of independent and homogeneous distribution often encounter a bottleneck in performance enhancement when processing pathology images due to neglecting the contextual relevance among patches. Some researchers have been keenly aware of this limitation and actively explored new strategies to integrate order-dependent and order-independent features, aiming to break through the performance limitations of the existing methods by enhancing the model's ability to capture complex spatial and structural information in pathology images\cite{shao2021transmil, chen2021multimodal, yang2024mambamil}. Among them, TransMIL\cite{shao2021transmil} is the first Transformer-based MIL method. It uses linear attention based on matrix decomposition\cite{xiong2021nystromformer} to capture order-independent features between instances, avoiding the high computational complexity problem of original attention\cite{vaswani2017attention} in WSIs. In addition, the use of order-dependent features is further enhanced by using the Pyramid Position Encoding Generator (PPEG), which consists of depthwise separable convolution at different scales. However, PPEG cannot fully satisfy the utilization of order-dependent features. MamMIL\cite{fang2024mammil} employs a structure similar to VisionMamba\cite{zhu2024vision}, modeling patch sequences at both forward and reverse levels, and integrates a PPEG-like multi-scale convolution. However, it does not analyze or exploit order-independent features. MambaMIL\cite{yang2024mambamil} is similar to MamMIL. The difference is that it adjusts the order of patch sequences to capture both order-dependent and order-independent features through a hyperparameter-controlled reordering method. Nonetheless, this reordering method is not adaptable to varying WSI sizes and may result in the loss of some local information. In this paper, we propose a sequence transformation method that addresses these limitations by accommodating WSIs with inconsistent sizes.

\section{Method}
\subsection{Overview}
Figure \ref{fig2} illustrates the structure of our architecture, which achieved the best results. Our framework supports the fusion of features from multiple branches and the following description uses three branches as an example. First, to exclude the influence of irrelevant regions, we discard the background. In large size WSIs (gigapixel images), not excluding the background causes a significant computational burden. Next, the WSIs are divided into patches, which are then embedded into feature vectors (also known as instance-level features) using a pretrained ResNet50. This process yields a sequence of instance features $S = \{x_1, x_2, x_3,\ldots, x_N\}$ of  , where $x_i \in \mathbb{R}^{D}$ denotes an instance feature, $N$ denotes length of sequence, and $D$ represents the feature dimension. The sequence $S$ is processed through feature reduction and squaring operations to produce $S'$, followed by sequence reordering to generate three distinct sequences: the original sequence $S_o$, the flipped sequence $S_f$ , and the transposed sequence $S_t$. These sequences are then efficiently modeled using stacked SSD blocks, resulting in the features $F_o$, $F_f$, and $F_t$ corresponding to each sequence, respectively. A feature weighted selection, implemented using the hyperbolic tangent activation function, is employed to select the features that contribute most effectively to the final result. Finally, the packet-level representation is derived using a Multi-Layer Perceptron (MLP) block. 
\subsection{Sequence squaring and reordering}
\begin{figure}[!t]
    \centering
    \includegraphics[width=0.6\linewidth]{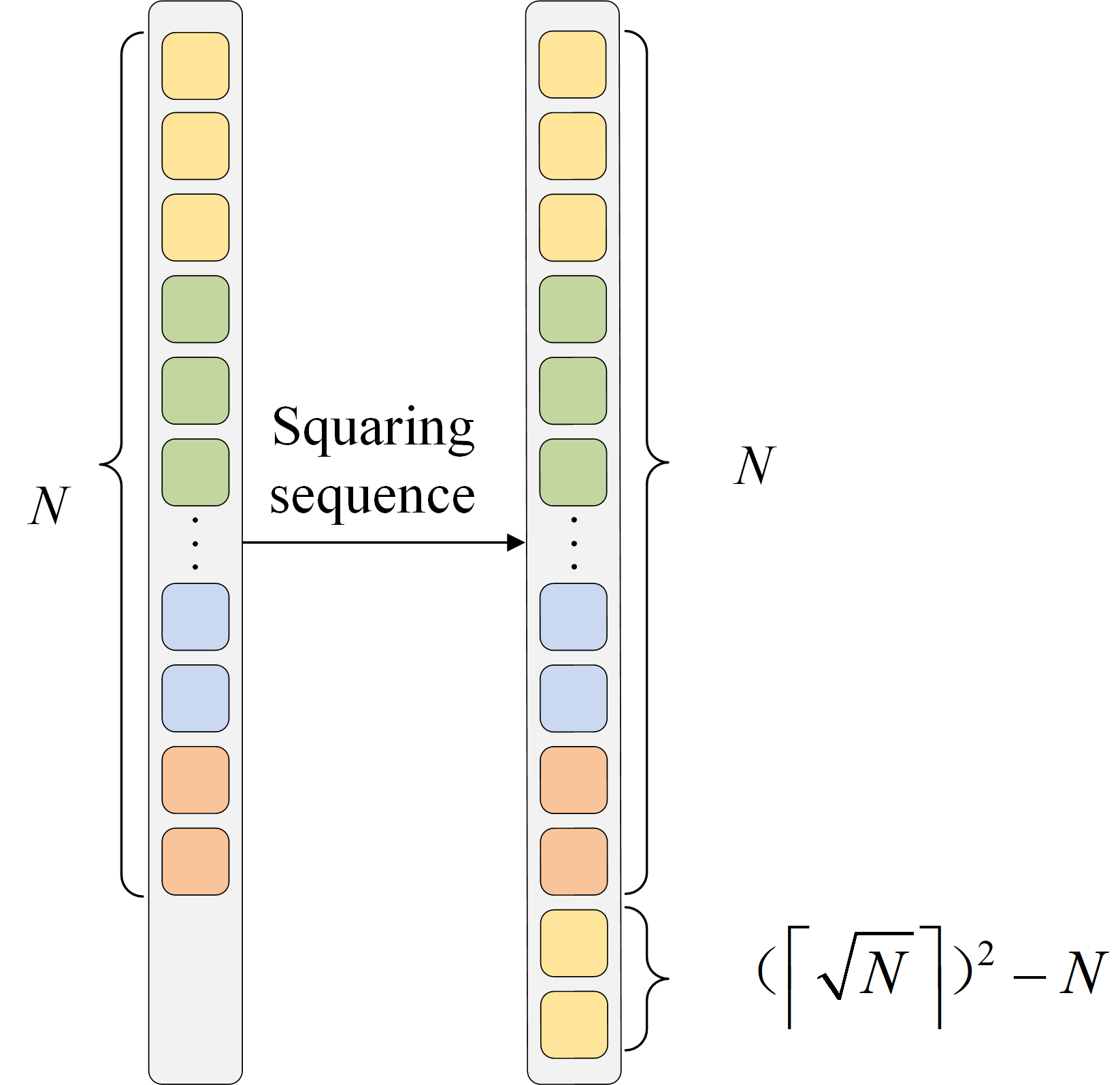}
    \caption{Square Sequence.}
    \label{fig3}
  \end{figure}

Given the large dimensionality (dimension = 1024) of the feature vectors embedded by the pretrained ResNet, we apply a linear projection to the sequence of instance features $S$ for dimensionality reduction, resulting in a reduced feature sequence $S_l$. To standardize the varying lengths of WSIs to a regular, manageable length, we then perform a squaring operation on $S_l$ to produce $S'$. As indicated in Table \ref{tab1}, the sizes of different WSIs can vary significantly, making it impractical to normalize them to a uniform size. The detailed sequence squaring process is illustrated in Figure \ref{fig3} and Equation \ref{eq1}. 
\begin{equation}
\begin{aligned}
& S^{\prime}=\operatorname{Concat}\left(S_l,\left(x_1, x_2, \ldots, x_M\right)\right) \\
& L=(\lceil\sqrt{N}\rceil)^2 \\
& M=L-N
\end{aligned}
\label{eq1}
\end{equation}
where $x_i$ denotes the strength feature, Concat denotes the feature splicing operation, $N$ denotes the original sequence length, $L$ denotes the sequence length after sequence squaring, and $M$ denotes the length of the padding in the sequence squaring operation. This method ensures that each WSI is processed appropriately while maintaining computational feasibility. After sequence squaring, the sequence of instance features can be represented as an $\lceil\sqrt{N}\rceil\times \lceil\sqrt{N}\rceil$ matrix, ready for sequence reordering.

\begin{figure}[!t]
    \centering
    \includegraphics[width=1.0\linewidth]{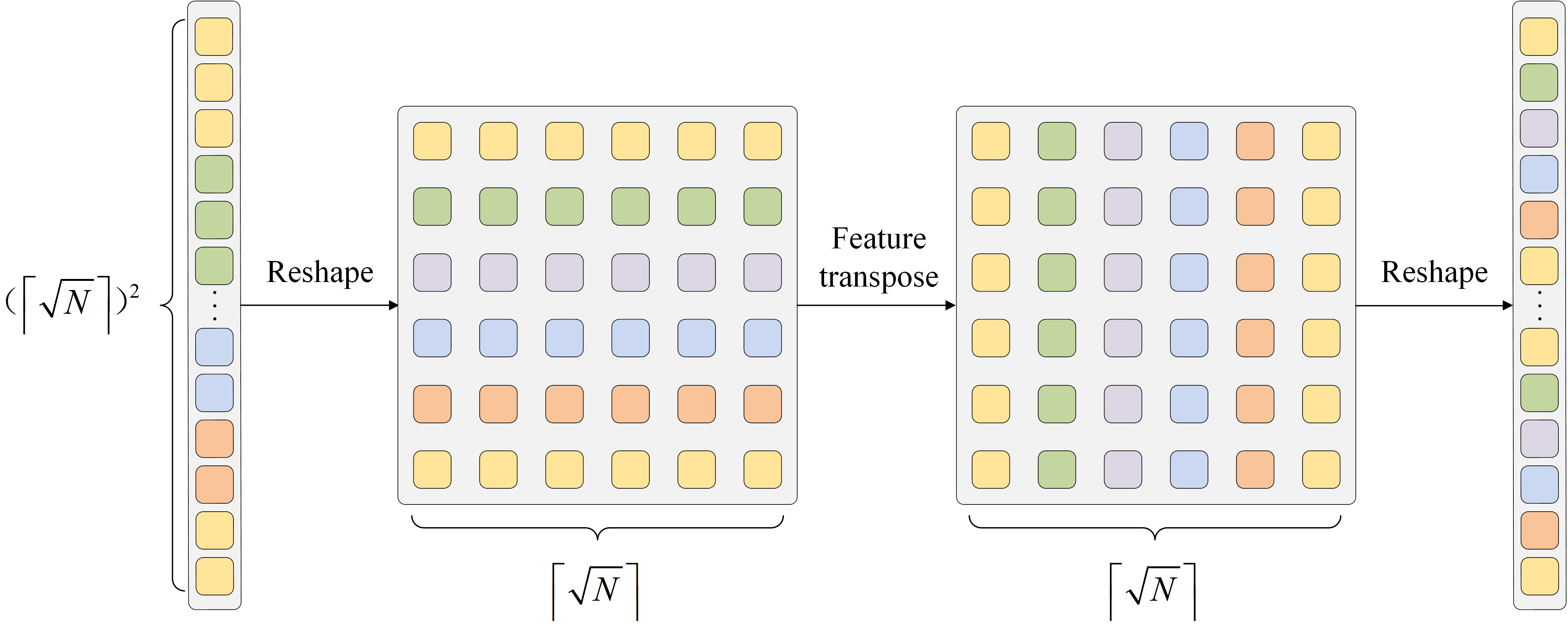}
    \caption{Transpose sequence.}
    \label{fig4}
  \end{figure}

  Sequence reordering in our approach involves two main operations: flip and transpose. The flip operation reverses the original sequence, addressing the history decay problem\cite{lu2024videomambapro} present in the original sequence within the SSD  block. The transpose operation, depicted in Figure \ref{fig4}, involves rearranging the sequence into a transposed sequence. Unlike the approach of Yang et al.\cite{yang2024mambamil}, we do not rely on fixed hyperparameters to disrupt the sequence structure. We argue that transposing the matrix $\lceil\sqrt{N}\rceil\times \lceil\sqrt{N}\rceil$, which is constructed based on the specific size of each WSI, is more suitable for handling the significant size variations across WSIs. Additionally, this method preserves the local information within the sequences and improves the utilization of sequence information, which can enhance the accuracy of classification. The original sequence $S_o$, the flipped sequence $S_f$, and the transposed sequence $S_t$ used as inputs for the SSD  block are defined in Equation \ref{eq2}.
\begin{equation}
\begin{aligned}
& S_o=S^{\prime} \\
& S_f=\operatorname{Flip}\left(S^{\prime}\right) \\
& S_t=\operatorname{Transpose}\left(S^{\prime}\right)
\end{aligned}
\label{eq2}
\end{equation}

\subsection{State space duality model}
The SSD algorithm is a more computationally efficient algorithm than previous SSM, building on SSM and starting with the same set of equations:
\begin{equation}
\begin{aligned}
h_t & =A_t h_{t-1}+B_t x_t \\
y_t & =C_t^{\top} h_t
\end{aligned}
\label{eq3}
\end{equation}
where $x_t$ is the input at moment $t$ and $y_t$ is the output at moment $t$. Equation \ref{eq3} defines a mapping from $x \in \mathbb{R}^{T}$ to $y \in \mathbb{R}^{T}$. We consider $x_t$ and $y_t$ as scalars and the hidden state $h_t$ as a one-dimensional vector of length $N$, where $N$ is called the state expansion factor and is an independent hyperparameter.

In SSM, the parameters of the three matrices $(A, B, C)$ can vary over time. Specifically, the tensor $A \in \mathbb{R}^{T \times N \times N)}$, while $B, C \in \mathbb{R}^{T \times N}$. To enhance computational efficiency, structured SSMs typically employ diagonal matrices for $A$. In such cases, $A$ effectively represents the diagonal elements of $N \times N$ matrix, allowing it to be simplified to $A \in \mathbb{R}^{T \times N}$.

The SSD algorithm in Mamba-2 introduces a further simplification by constraining the diagonal matrix $A$ such that all its diagonal elements are identical, and all other elements are zero. Consequently, $A$ can be represented as a scalar, or it can be parameterized by $T$ alone. This simplification greatly reduces computational complexity  and allows for the combination of SSM and attention while maintaining the structural integrity of the model. In this paper, we use Equation \ref{eq4} to represent the entire SSD process.
\begin{equation}
\begin{aligned}
& F=\text {Linear}(\operatorname{SSD}(\operatorname{Conv1D}(\text{Linear}(S)))) \\
& F \in\left\{F_o, F_f, F_t\right\}, S \in\left\{S_o, S_f, S_t\right\}
\end{aligned}
\label{eq4}
\end{equation}
where $F, S \in \mathbb{R}^{L \times D'}$, $D'$ denotes the dimension of $D$ after dimensionality reduction.

\begin{figure*}[!t]
    \centering
    \begin{minipage}[t]{0.8\linewidth}
    \centering
        \begin{tabular}{@{\extracolsep{\fill}}c@{}c@{}c@{}@{\extracolsep{\fill}}}
            \includegraphics[width=0.33\linewidth]{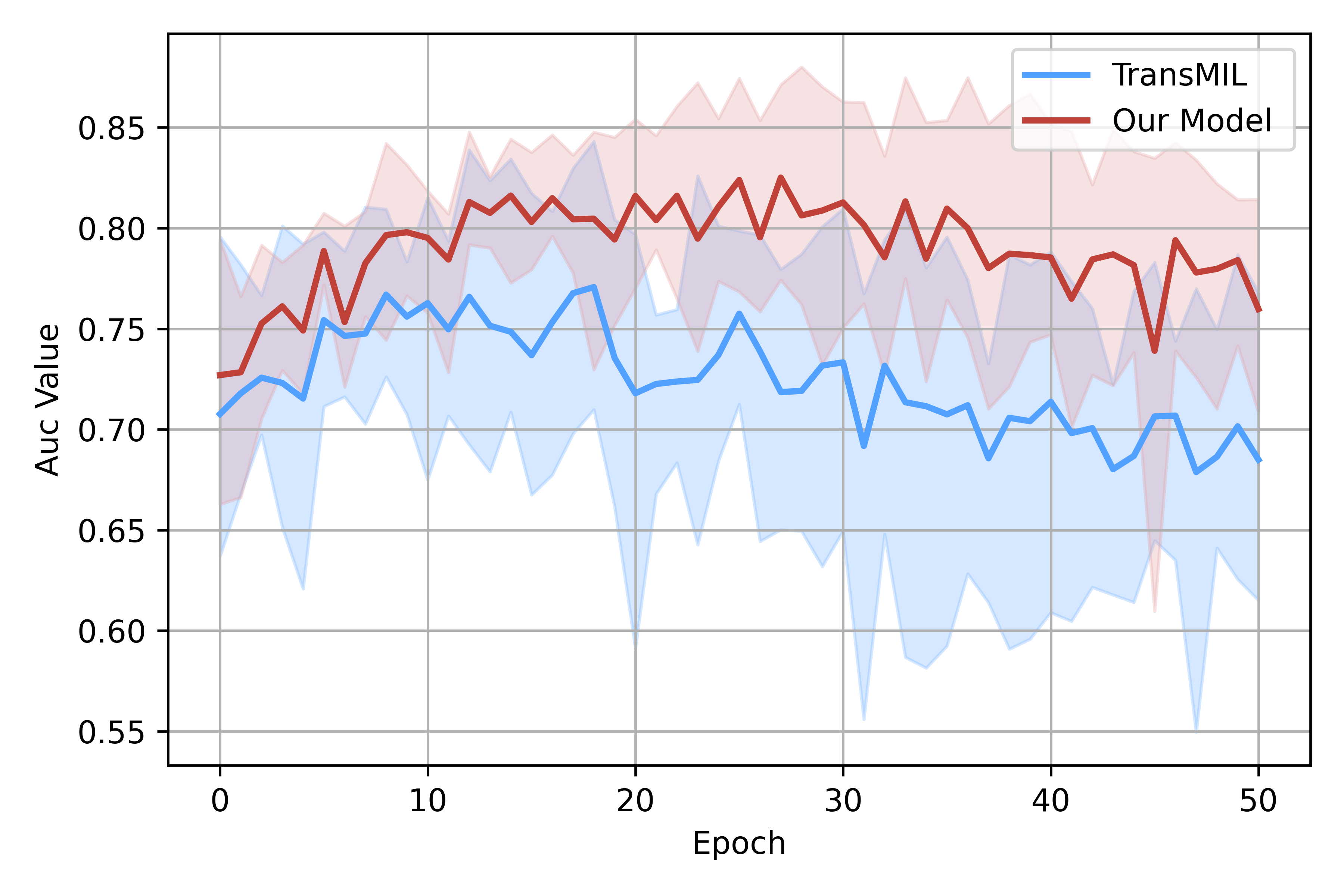} &
            \includegraphics[width=0.33\linewidth]{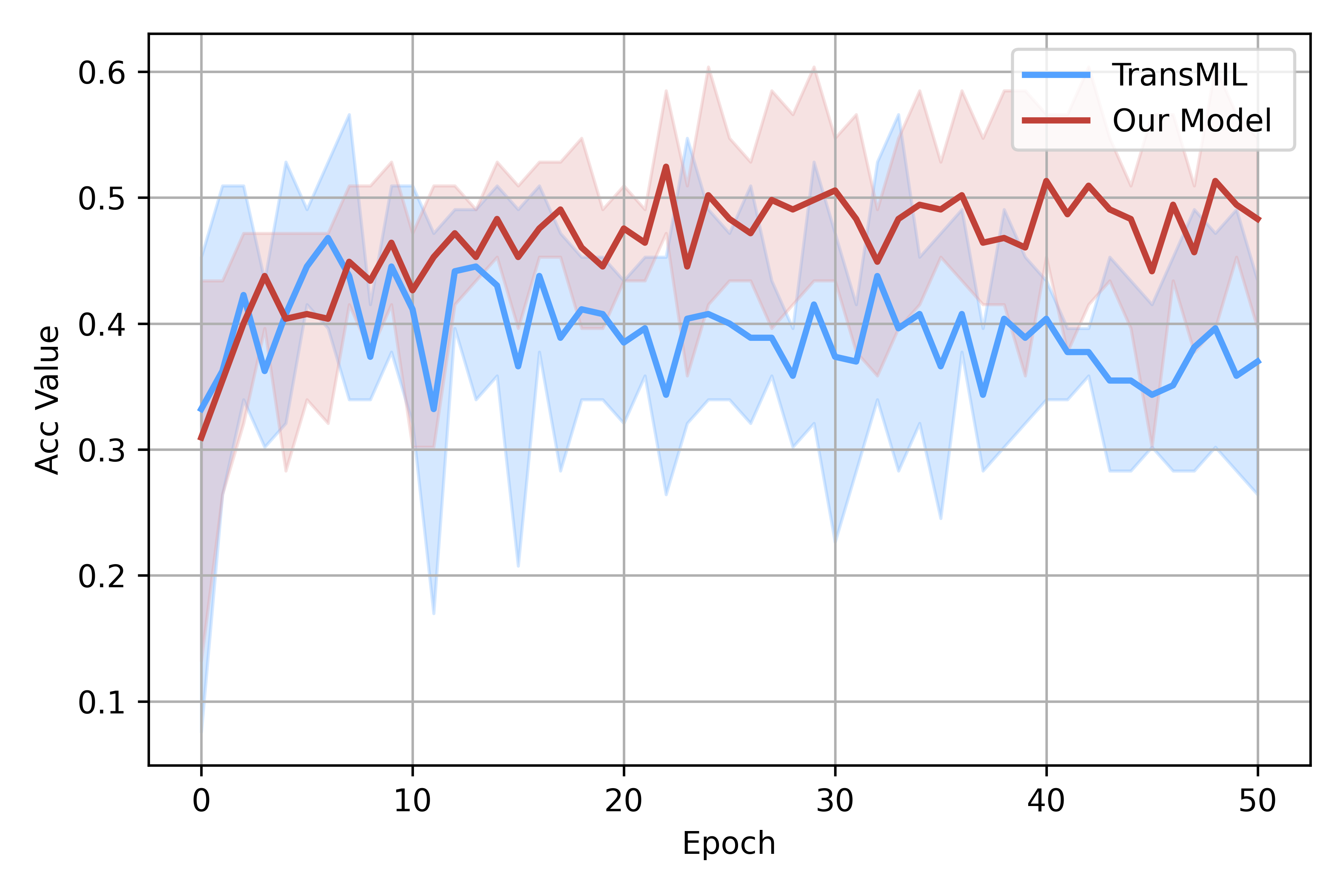} &
            \includegraphics[width=0.33\linewidth]{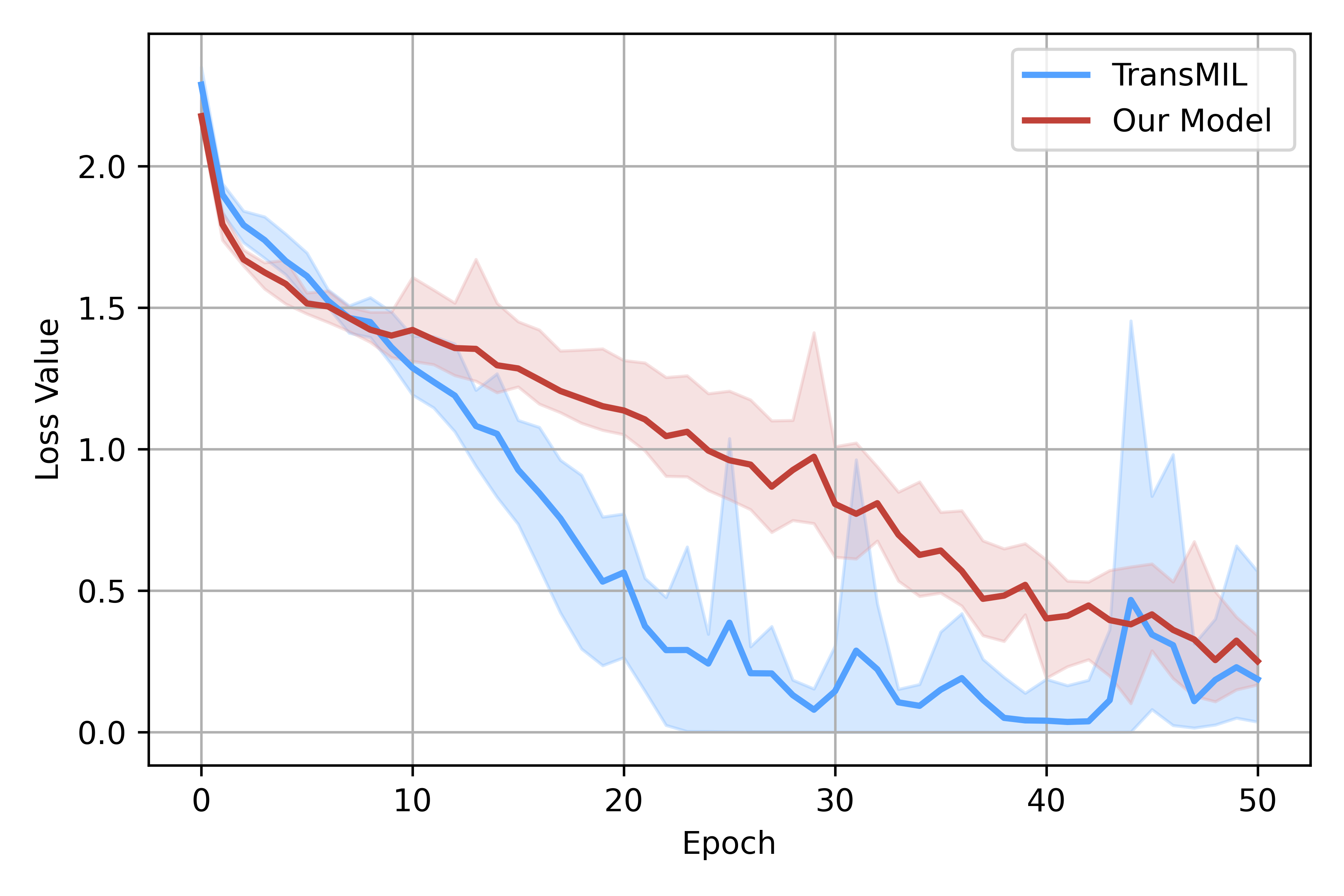}\\
            \footnotesize(a) & \footnotesize(b) & \footnotesize(c)\\
        \end{tabular}
    \end{minipage}
    \begin{minipage}[t]{0.8\linewidth}
    \centering
        \begin{tabular}{@{\extracolsep{\fill}}c@{}c@{}c@{}@{\extracolsep{\fill}}}
            \includegraphics[width=0.33\linewidth]{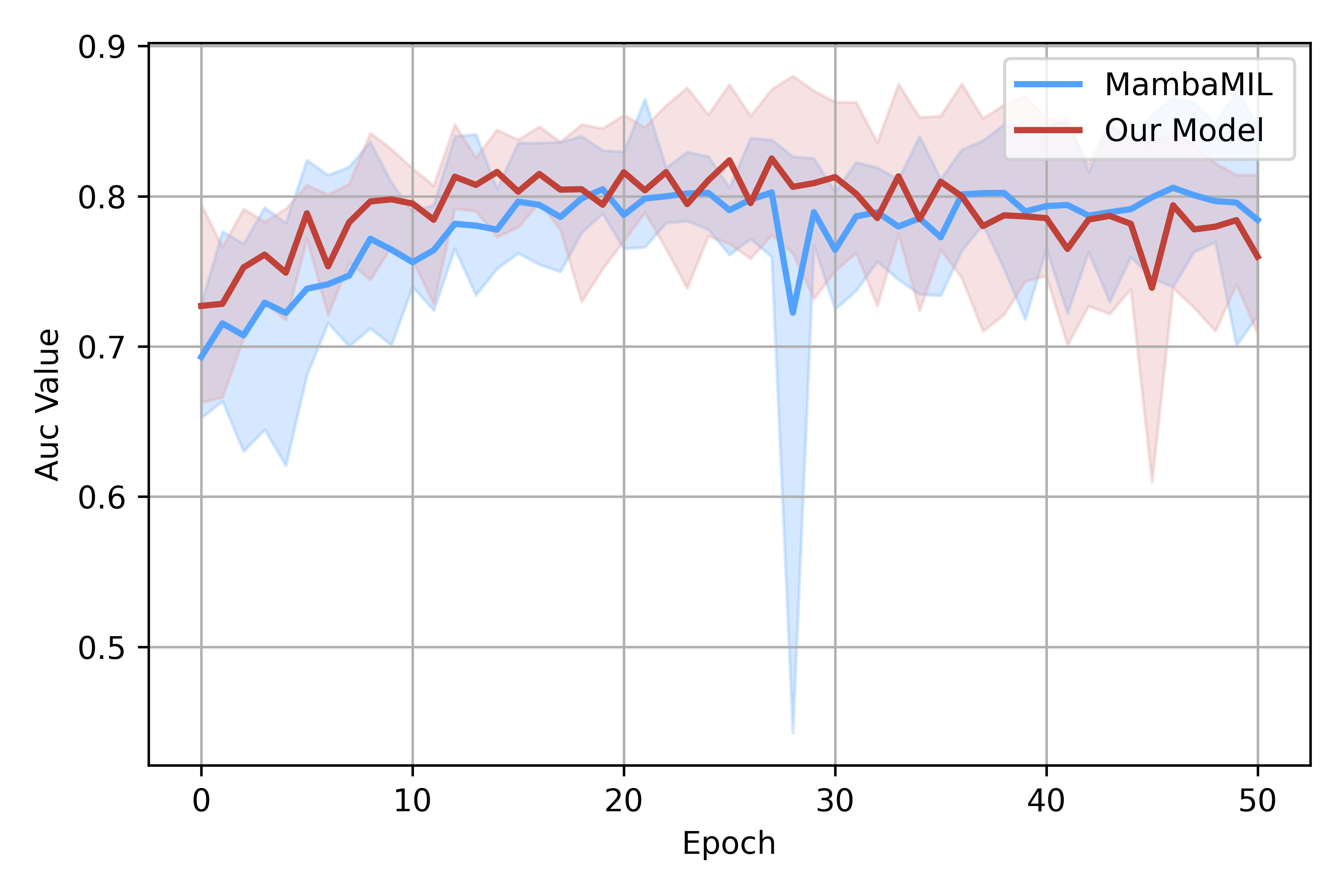} &
            \includegraphics[width=0.33\linewidth]{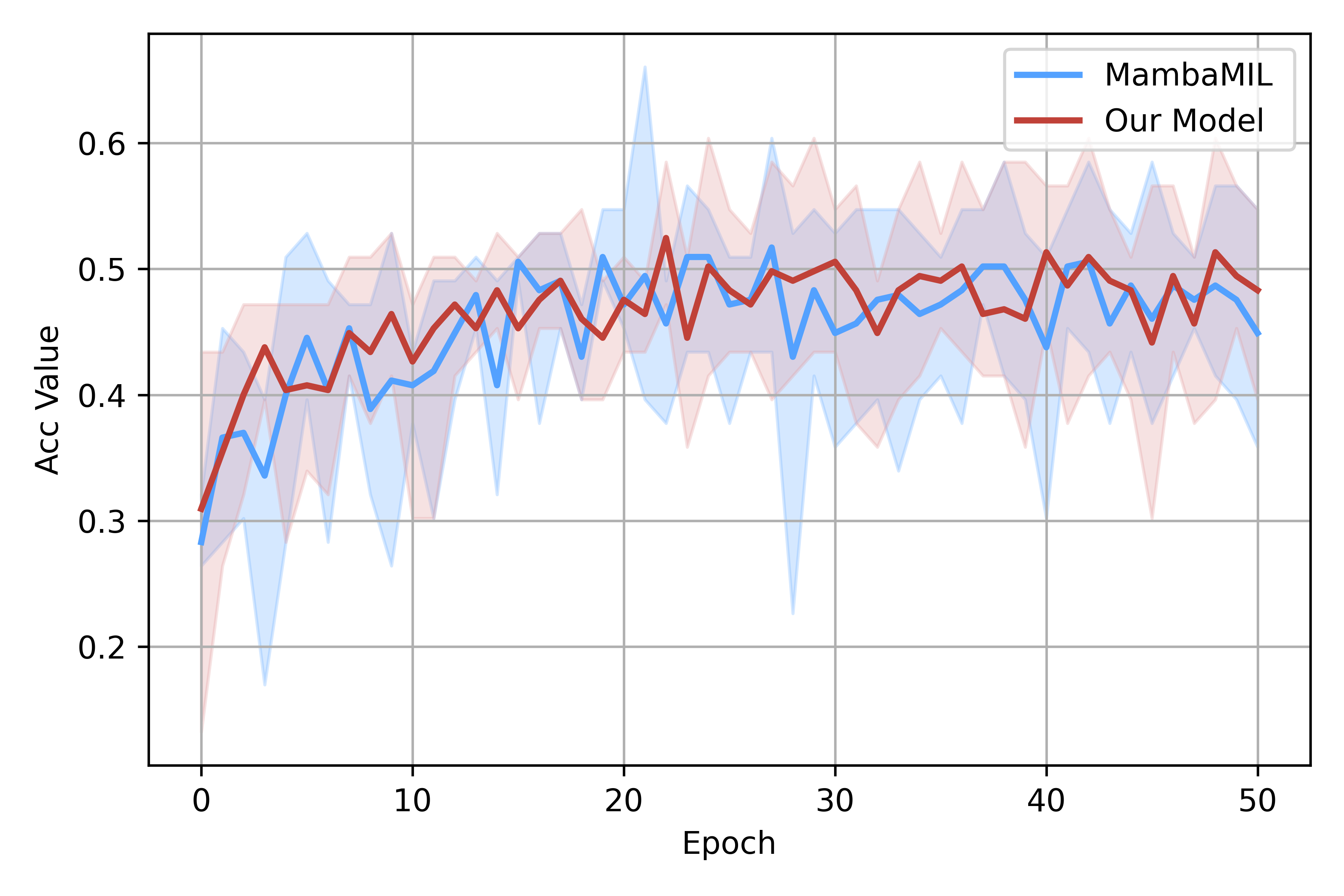} &
            \includegraphics[width=0.33\linewidth]{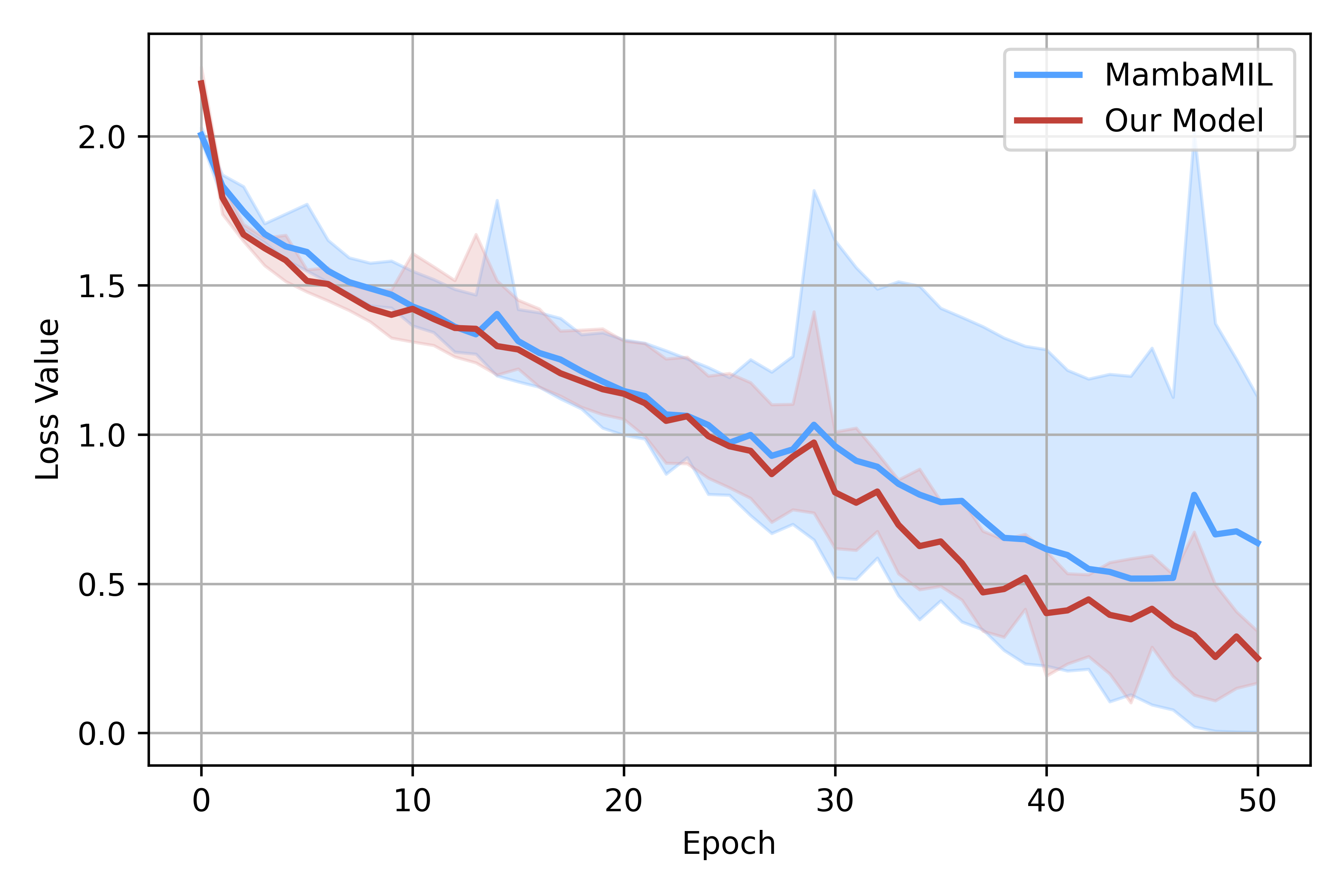}\\
            \footnotesize(d) & \footnotesize(e) & \footnotesize(f)\\
        \end{tabular}
    \end{minipage}
    \caption{The performance of our model and other models in the validation set on BRCAS dataset. (a-c)Comparison results between our model and TransMIL. (d-f)Comparison results between our model and MambaMIL.}
    \label{fig5}
 \end{figure*}

\subsection{Feature selection}
After splicing the features that capture different sequences using the stacked SSD  block, we use the hyperbolic tangent activation function to achieve feature weighted selection. Specifically, as shown in Equation \ref{eq5}, the result of the linear transformation is nonlinearly activated by applying the hyperbolic tangent activation function (Tanh) through a linear layer that first maps the input features to a lower dimensional space. Subsequently, the Tanh-activated features are mapped into weight vectors via an additional linear layer. These weights are normalized by a softmax function and multiplied with the spliced features to obtain a weighted feature representation. This approach allows the model to automatically adjust the level of attention to individual features depending on the input data.
\begin{equation}
\begin{aligned}
& F_c=\text {Concat}(F_o, F_f, F_t) \\
& F_c'=\text {Linear}(\operatorname{Tanh}(\text{Linear}(F_c))) \\
& R=F_c \times \text {Softmax}(F_c')
\end{aligned}
\label{eq5}
\end{equation}
where $F_c$ denotes the spliced features, $F_c'$ denotes the weight vector, and $R$ denotes the result after feature weighted selection.

\begin{table}[htbp]
    \caption{Sample information for BRACS dataset and NSCLC dataset}
    \label{tab1}
    \begin{center}
    \begin{tabular}{@{}lll@{}}
    \toprule
                     & BRACS        & NSCLC        \\ \midrule
    Sample           & 537          & 1052         \\
    Min Image Size         & 23799$\times$15455  & 10000$\times$4617   \\
    Max Image Size         & 181272$\times$88334 & 191352$\times$97078 \\
    Min Bag Size     & 46           & 41           \\
    Max Bag Size     & 7728         & 11019        \\
    Average Bag Size & 2587         & 3004         \\ \bottomrule
    \end{tabular}
    \end{center}
    \end{table}

\section{Experiments}
\subsection{Datasets}
To validate the effectiveness of our model, we conducted experiments using two datasets: BRACS\cite{brancati2022bracs} and NSCLC. The sample details of datasets are summarized in Table \ref{tab1}.

\textbf{BRACS Dataset.} BRACS is a histopathology image dataset of hematoxylin and eosin (H\&E) stained breast cancer tissues. As shown in Table \ref{tab2}, it includes a total of seven subtypes: normal (glandular tissue samples without lesions), pathologically benign (PB), usual ductal hyperplasia (UDH), flat epithelial atypia (FEA), atypical ductal hyperplasia (ADH), ductal carcinoma in situ (DCIS), and invasive carcinoma (IC). The dataset comprises 537 samples, with image sizes ranging from a maximum of 181,272 $\times$ 88,334 pixels to a minimum of 23,799 $\times$ 15,455 pixels. The average sample size is approximately 2,587 patches.

\begin{table}[htbp]
    \caption{BRACS dataset subtypes distribution}
    \label{tab2}
    \begin{center}
    \begin{tabular}{@{}ll@{}}
    \toprule
    Subtypes                    & Number \\ \midrule
    Normal                      & 40     \\
    Pathological Benign         & 145    \\
    Usual Ductal Hyperplasia    & 70     \\
    Flat   Epithelial Atypia    & 41     \\
    Atypical Ductal Hyperplasia & 48     \\
    Ductal Carcinoma In Situ    & 61     \\
    Invasive Carcinoma          & 132    \\ \bottomrule
    \end{tabular}
    \end{center}
    \end{table}

\textbf{NSCLC Dataset.} NSCLC, sourced from The Cancer Genome Atlas (TCGA), consists of H\&E-stained histopathology images of lung squamous cell carcinoma and lung adenocarcinoma. It includes two subtypes: lung squamous cell carcinoma (LUSC) and lung adenocarcinoma (LUAD). The dataset comprises 537 samples, with image sizes ranging from a maximum of 191,352 $\times$ 97,078 pixels to a minimum of 10,000 $\times$ 4,617 pixels. The average sample size is approximately 3,004 patches.

\begin{table*}[t!]
    \caption{Results of WSIs classification}
    \label{tab3}
    \begin{center}
    \resizebox{\linewidth}{!}{
    \begin{tabular}{@{}lllll|llll|llll@{}}
    \toprule
                 & \multicolumn{4}{c|}{BRACS}                                    & \multicolumn{4}{c|}{NSCLC}                                    & \multicolumn{4}{c}{MEAN}                \\ \midrule
                 & Test AUC      & Test ACC      & Val AUC       & Val ACC       & Test AUC      & Test ACC      & Val AUC       & Val ACC       & Test AUC & Test ACC & Val AUC & Val ACC \\ \midrule
    Max Pooling  & 0.7470\pmtext{0.0384} & 0.4189\pmtext{0.0430} & 0.7894\pmtext{0.0083} & 0.4792\pmtext{0.0286} & 0.9395\pmtext{0.0076} & \underline{0.8788\pmtext{0.0264}} & 0.9142\pmtext{0.0396} & 0.8373\pmtext{0.0495} & 0.8433   & 0.6489   & 0.8518  & 0.6583  \\
    Mean Pooling & 0.7659\pmtext{0.0440} & 0.4717\pmtext{0.0611} & 0.7987\pmtext{0.0379} & 0.4717\pmtext{0.0626} & 0.9042\pmtext{0.0111} & 0.8277\pmtext{0.0196} & 0.8677\pmtext{0.0333} & 0.7943\pmtext{0.0542} & 0.8351   & 0.6497   & 0.8332  & 0.6330  \\
    ABMIL\cite{ilse2018attention}        & 0.7872\pmtext{0.0271} & \textbf{0.5094\pmtext{0.0566}} & 0.7944\pmtext{0.0612} & 0.5208\pmtext{0.0930} & 0.9381\pmtext{0.0144} & 0.8630\pmtext{0.0124} & 0.8971\pmtext{0.0449} & 0.8337\pmtext{0.0317} & 0.8627   & \underline{0.6862}   & 0.8458  & 0.6773  \\
    CLAM-SB \cite{lu2021data}        & 0.7885\pmtext{0.0463} & 0.4415\pmtext{0.0606} & 0.8154\pmtext{0.0363} & 0.4604\pmtext{0.0492} & \underline{0.9491\pmtext{0.0139}} & 0.8727\pmtext{0.0166} & 0.9147\pmtext{0.0435} & 0.8396\pmtext{0.0327} & \underline{0.8688}   & 0.6571   & 0.8651  & 0.6500  \\
    TransMIL \cite{shao2021transmil}    & 0.7333\pmtext{0.0415} & 0.4226\pmtext{0.0434} & 0.7951\pmtext{0.0456} & 0.4566\pmtext{0.0506} & 0.9159\pmtext{0.0246} & 0.8310\pmtext{0.0535} & 0.8923\pmtext{0.0502} & 0.8135\pmtext{0.0444} & 0.8246   & 0.6268   & 0.8437  & 0.6351  \\
    S4MIL\cite{fillioux2023structured}        & 0.7816\pmtext{0.0260} & 0.4792\pmtext{0.0544} & 0.8196\pmtext{0.0477} & \underline{0.5283\pmtext{0.0267}} & 0.8865\pmtext{0.0702} & 0.7937\pmtext{0.0550} & 0.8523\pmtext{0.0697} & 0.7590\pmtext{0.0771} & 0.8341   & 0.6365   & 0.8360  & 0.6437  \\
    MambaMIL\cite{yang2024mambamil}     & \underline{0.7912\pmtext{0.0261}} & 0.4604\pmtext{0.0560} & \underline{0.8233\pmtext{0.0192}} & 0.5208\pmtext{0.0286} & 0.9315\pmtext{0.0204} & 0.8580\pmtext{0.0389} & \underline{0.9175\pmtext{0.0216}} & \underline{0.8413\pmtext{0.0232}} & 0.8614   & 0.6592   & \underline{0.8704}  & \underline{0.6811}  \\
    Our Model    & \textbf{0.7986\pmtext{0.0340}} & \underline{0.4981\pmtext{0.0510}} & \textbf{0.8364\pmtext{0.0203}} & \textbf{0.5321\pmtext{0.0310}} & \textbf{0.9533\pmtext{0.0215}} & \textbf{0.8794\pmtext{0.0348}} & \textbf{0.9202\pmtext{0.0436}} & \textbf{0.8491\pmtext{0.0511}} & \textbf{0.8760}   & \textbf{0.6888}   & \textbf{0.8783}  & \textbf{0.6906}  \\ \bottomrule
    \end{tabular}}
    \end{center}
    \end{table*}

\subsection{Implementation Details}

Each WSI is divided into a series of non-overlapping 512$\times$512 patches at $\times$20 magnification, and the background regions are discarded. ResNet\cite{he2016deep}, pre-trained on ImageNet\cite{deng2009imagenet}, is used to embed the features of each patch into a 1024-dimensional vector. A linear projection layer is then applied to reduce these embeddings from 1024 to 512 dimensions. The features of each bag are thus represented as a vector in $B_i \in \mathbb{R}^{L \times 512}$. To mitigate the impact of data partitioning and randomness on model evaluation, we utilized 5-fold cross-validation across both datasets. The datasets were split into training, validation, and test sets in an 8:1:1 ratio. Consistent with standard practices, we employed the area under the curve (AUC) and accuracy (ACC) metrics, along with their standard deviations (std), for evaluation. In all tables, the best result is indicated in bold, while the second-best result is underlined.

\textbf{Baseline.} We selected several methods: (1) traditional pooling methods, including average pooling and maximum pooling; (2) the attention-based pooling method ABMIL\cite{ilse2018attention}; (3) the single-attention branching model CLAM-SB\cite{lu2021data}; (4) TransMIL\cite{shao2021transmil}, which incorporates linear attention and positional encoding; (5) the SSM-based model S4MIL\cite{fillioux2023structured}; and (6) MambaMIL\cite{yang2024mambamil}, which integrates SSM and sequence transformation techniques. These baselines provide a range of approaches for aggregating features and serve as benchmarks to evaluate the performance of our proposed method. We use the same settings to ensure that a fair comparison is achieved. During training, cross-entropy loss is used as the loss function, and the Adam optimizer is employed with a learning rate of $2 \times 10^{-4}$ and a batch size of 1.

\begin{figure}[!t]
    \centering
    \includegraphics[width=1.0\linewidth]{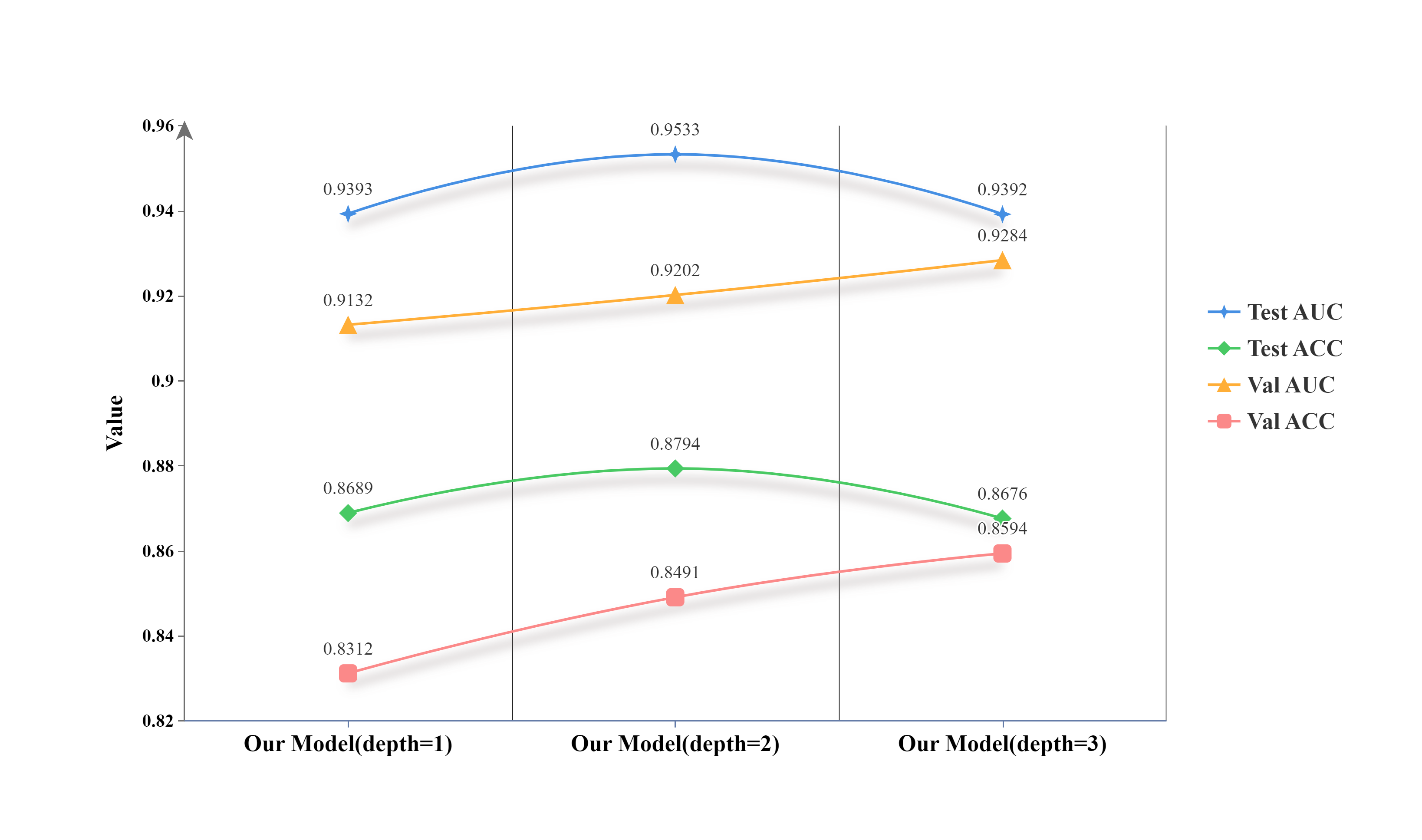}
    \caption{Results of our model with different depths on the NSCLC dataset}
    \label{fig6}
  \end{figure}

  \begin{table}[htbp]
    \caption{Comparison results between our model and Mamba-2 on the NSCLC dataset}
    \label{tab4}
    \begin{center}
    \resizebox{\linewidth}{!}{
    \begin{tabular}{@{}lllll@{}}
    \toprule
                      & Test AUC      & Test ACC      & Val AUC       & Val ACC       \\ \midrule
    Mamba-2          & \underline{0.9479\pmtext{0.0191}} & \underline{0.8710\pmtext{0.0232}} & 0.9170\pmtext{0.0296} & 0.8274\pmtext{0.0274} \\
    Mamba-2(wrapping) & 0.9335\pmtext{0.0227} & 0.8609\pmtext{0.0217} & \underline{0.9199\pmtext{0.0159}} & \underline{0.8364\pmtext{0.0274}} \\
    Our Model         & \textbf{0.9533\pmtext{0.0215}} & \textbf{0.8794\pmtext{0.0348}} & \textbf{0.9202\pmtext{0.0436}} & \textbf{0.8491\pmtext{0.0511}} \\ \bottomrule
    \end{tabular}}
    \end{center}
    \end{table}

\subsection{Results on WSIs classification}
Table \ref{tab3} presents the experimental results for the BRACS and NSCLC datasets. Our proposed method, Mamba2MIL, achieved near-optimal performance across almost all metrics. On the BRACS test set, the performance of our model was only marginally lower than ABMIL, with a difference of approximately 0.01 in ACC. Specifically, our model achieved an AUC of 0.7986 and an ACC of 0.4981 on the BRACS dataset, while on the NSCLC dataset, it achieved an AUC of 0.9533 and an ACC of 0.8794. Compared to MambaMIL, our model showed an AUC improvement of approximately 0.01 on the BRACS dataset and an ACC improvement of around 0.04. On the NSCLC dataset, the AUC improvement was smaller, but the ACC improved by about 0.02.

As illustrated in Figure \ref{fig5}, we further visualized the performance differences between our model and others on the BRACS validation set. Figure \ref{fig5}-(a-c) demonstrate that our proposed method outperforms TransMIL in the validation set AUC and ACC during 5-fold cross-validation, with more stable loss behavior, although the loss reduction was not as rapid as in TransMIL. Additionally, as shown in Figure \ref{fig5}-(d-e), while the performance of our proposed method closely matched that of MambaMIL in terms of the validation set AUC and ACC, it performed better on the test set, suggesting stronger generalization capability. Figure \ref{fig5}-f indicates that the loss reduction in our proposed model is both faster and more stable compared to that of MambaMIL, highlighting the efficiency of our approach.

\begin{table*}[htbp]
    \caption{Results of our model on the NSCLC dataset with different settings}
    \label{tab8}
    \begin{center}
    \resizebox{\linewidth}{!}{
    \begin{tabular}{@{}llllllll|llll@{}}
    \toprule
    Mamba-2 & Mamba & Original order & flipped order & transposed order & Add & \multicolumn{2}{l|}{Concatenate} & Test AUC      & Test ACC      & Val AUC       & Val ACC       \\ \midrule
    $\checkmark$       &       & $\checkmark$              & $\checkmark$          & $\checkmark$               &     & $\checkmark$               &                & \underline{0.9533\pmtext{0.0215}} & 0.8794\pmtext{0.0348} & 0.9202\pmtext{0.0436} & \textbf{0.8491\pmtext{0.0511}} \\
    $\checkmark$       &       & $\checkmark$              & $\checkmark$          & $\checkmark$               & $\checkmark$   &                 &                & \textbf{0.9535\pmtext{0.0095}} & \underline{0.8839\pmtext{0.0173}} & 0.9139\pmtext{0.0428} & \underline{0.8416\pmtext{0.0407}} \\
            & $\checkmark$     & $\checkmark$              & $\checkmark$          & $\checkmark$               &     & $\checkmark$               &                & 0.9397\pmtext{0.0138} & \textbf{0.8866\pmtext{0.0166}} & \underline{0.9243\pmtext{0.0259}} & 0.8409\pmtext{0.0376} \\
    $\checkmark$       &       & $\checkmark$              & $\checkmark$          &                 &     & $\checkmark$               &                & 0.9403\pmtext{0.0269} & 0.8685\pmtext{0.0118} & 0.9049\pmtext{0.0349} & 0.8207\pmtext{0.0475} \\
    $\checkmark$       &       & $\checkmark$              &            & $\checkmark$               &     & $\checkmark$               &                & 0.9442\pmtext{0.0026} & 0.8642\pmtext{0.0488} & 0.9137\pmtext{0.0457} & 0.8374\pmtext{0.0551} \\
    $\checkmark$       &       &                & $\checkmark$          & $\checkmark$               &     & $\checkmark$               &                & 0.9489\pmtext{0.0074} & 0.8726\pmtext{0.0102} & 0.9210\pmtext{0.0383} & 0.8299\pmtext{0.0433} \\
    $\checkmark$       &       & $\checkmark$              &            &                 &     &                 &                & 0.9479\pmtext{0.0191} & 0.8710\pmtext{0.0232} & 0.9170\pmtext{0.0296} & 0.8274\pmtext{0.0274} \\
    $\checkmark$       &       &                & $\checkmark$          &                 &     &                 &                & 0.9429\pmtext{0.0093} & 0.8667\pmtext{0.0191} & \textbf{0.9246\pmtext{0.0419}} & 0.8396\pmtext{0.0412} \\
    $\checkmark$       &       &                &            & $\checkmark$               &     &                 &                & 0.9276\pmtext{0.0577} & 0.8839\pmtext{0.0494} & 0.9057\pmtext{0.0366} & 0.8244\pmtext{0.0379} \\ \bottomrule
    \end{tabular}}
    \end{center}
    \end{table*}

\subsection{Comparison Results}
To thoroughly evaluate the effectiveness of Mamba2MIL, we conducted extensive comparative experiments against other models. As shown in Table \ref{tab4}, our model achieved the best results across all metrics compared to the two variants of Mamba-2, demonstrating the efficacy of our improvements. In these experiments, Mamba-2 (wrapping) included an additional mixer class and residual connections to Mamba-2.

Figure \ref{fig6} presents the performance of our model at different depths, with the best results obtained at a depth of two layers. While a three-layer depth showed slightly better performance on the validation set, it performed poorly on the test set, indicating a lack of generalization. Table \ref{tab6} outlines the results of comparative experiments concerning a critical parameter of the Mamba model—the state expansion factor. The optimal performance was achieved with a state expansion factor of 64. Although Mamba-2 supports larger state dimensions than Mamba-1, this does not necessarily mean that the largest state dimensions are optimal for WSIs classification.

Additionally, we conducted experiments under the same settings but replaced the transpose order method with other reordering methods for comparison. The results, presented in Table \ref{tab7}, indicate that the transpose order method outperforms the reordering method used in MambaMIL. Although the randomized reordering method showed better results on the NSCLC dataset, it performed worse on the BRACS dataset, with AUC and ACC scores of 0.7841 and 0.4603 on the test set, and 0.8304 and 0.5509 on the validation set, respectively. This suggests that while randomized reordering can help the model capture more order-independent features for certain datasets, it may compromise generalization performance. In contrast, the transposed sequence method demonstrated consistently excellent performance across datasets, suggesting it aids the model in capturing more useful features with robust generalization capabilities.

\begin{table}[htbp]
    \caption{Comparison results of different state expansion factors for our model on the NSCLC dataset}
    \label{tab6}
    \begin{center}
    \resizebox{\linewidth}{!}{
    \begin{tabular}{@{}lllll@{}}
    \toprule
                            & Test AUC      & Test ACC      & Val AUC       & Val ACC       \\ \midrule
    Our Model+LN            & 0.9353\pmtext{0.0236} & 0.8653\pmtext{0.0338} & 0.9176\pmtext{0.0340} & 0.8483\pmtext{0.0277} \\
    Our Model(state=64)     & \textbf{0.9533\pmtext{0.0215}} & \textbf{0.8794\pmtext{0.0348}} & \underline{0.9202\pmtext{0.0436}} & \underline{0.8491\pmtext{0.0511}} \\
    Our Model(state=128)    & \underline{0.9421\pmtext{0.0166}} & \underline{0.8711\pmtext{0.0242}} & \textbf{0.9228\pmtext{0.0298}} & \textbf{0.8500\pmtext{0.0327}} \\ \bottomrule
    \end{tabular}}
    \end{center}
    \end{table}

\subsection{Ablation Study}
To further assess the impact of multiple reordering methods on model performance, we conducted a series of ablation studies. As shown in Table 8, the first three rows encompass almost all of the best and second-best results. These experiments all use three orders of features, differing only slightly in the SSD block and feature aggregation steps. Specifically, the 'Add' and 'Concatenate' operations indicate different ways of aggregating the results of the SSD  block, respectively. While the summing-based feature aggregation method appears to outperform Mamba2MIL, similar to random reordering, it struggles on the BRACS dataset (Test AUC=0.7751, Test ACC=0.4302)and shows limited generalization. Incorporating the transposition order improves the test set AUC and ACC from 0.9403 and 0.8685 to 0.9533 and 0.8794, respectively. Similarly, adding the flip order enhances the test set AUC and ACC from 0.9442 and 0.8642 to 0.9533 and 0.8794, respectively. These results suggest that sequence augmentation methods that aggregate multiple alignments are capable of extracting more discriminative representations from various sequence orders, thereby effectively enhancing WSI classification performance.

\begin{table}[htbp]
    \caption{Comparison results of different ordering methods for our model on the NSCLC dataset}
    \label{tab7}
    \begin{center}
    \resizebox{\linewidth}{!}{
    \begin{tabular}{@{}lllll@{}}
    \toprule
                            & Test AUC      & Test ACC      & Val AUC       & Val ACC       \\ \midrule
    Our Model     & \underline{0.9533\pmtext{0.0215}} & \textbf{0.8794\pmtext{0.0348}} & 0.9202\pmtext{0.0436} & \underline{0.8491\pmtext{0.0511}} \\
    Our Model(MambaMIL reorder)      & 0.9493\pmtext{0.0138} & \underline{0.8789\pmtext{0.0233}} & \underline{0.9210\pmtext{0.0467}} & 0.8325\pmtext{0.0416} \\
    Our Model(random order) & \textbf{0.9556\pmtext{0.0135}} & 0.8764\pmtext{0.0129} & \textbf{0.9327\pmtext{0.0363}} & \textbf{0.8535\pmtext{0.0410}} \\ \bottomrule
    \end{tabular}}
    \end{center}
    \end{table}

\section{Conclusion}
In this paper, we propose a novel MIL framework called Mamba2MIL. Our framework utilizes SSD to model long sequences of patches of WSIs, which, combined with weighted feature selection, supports the fusion processing of more branching features and can be extended according to specific application needs. Moreover, we introduce a sequence transformation method tailored to varying WSI sizes, which enhances sequence-independent features while preserving local sequence information, thereby improving sequence information utilization. We evaluate our model against state-of-the-art methods using 5-fold cross-validation on two datasets with different cancer types, achieving consistently optimal performance. These results validate the effectiveness of Mamba2MIL in enhancing the accuracy and reliability of computational pathology, underscoring its potential as a valuable tool for pathologists in clinical practice.




\section*{Acknowledgment}

This study is partially supported by National Natural Science Foundation of China (62176016, 72274127), National Key R\&D Program of China (No. 2021YFB2104800), Guizhou Province Science and Technology Project: Research on Q\&A Interactive Virtual Digital People for Intelligent Medical Treatment in Information Innovation Environment (supported by Qiankehe[2024] General 058), Capital Health Development Research Project(2022-2-2013), Haidian innovation and translation program from Peking University Third Hospital (HDCXZHKC2023203), and Project: Research on the Decision Support System for Urban and Park Carbon Emissions Empowered by Digital Technology - A Special Study on the Monitoring and Identification of Heavy Truck Beidou Carbon Emission Reductions.

\bibliographystyle{IEEEtran}
\bibliography{references}

\begin{thebibliography}{10}
\providecommand{\url}[1]{#1}
\csname url@samestyle\endcsname
\providecommand{\newblock}{\relax}
\providecommand{\bibinfo}[2]{#2}
\providecommand{\BIBentrySTDinterwordspacing}{\spaceskip=0pt\relax}
\providecommand{\BIBentryALTinterwordstretchfactor}{4}
\providecommand{\BIBentryALTinterwordspacing}{\spaceskip=\fontdimen2\font plus
\BIBentryALTinterwordstretchfactor\fontdimen3\font minus \fontdimen4\font\relax}
\providecommand{\BIBforeignlanguage}[2]{{%
\expandafter\ifx\csname l@#1\endcsname\relax
\typeout{** WARNING: IEEEtran.bst: No hyphenation pattern has been}%
\typeout{** loaded for the language `#1'. Using the pattern for}%
\typeout{** the default language instead.}%
\else
\language=\csname l@#1\endcsname
\fi
#2}}
\providecommand{\BIBdecl}{\relax}
\BIBdecl

\bibitem{chen2024towards}
R.~J. Chen, T.~Ding, M.~Y. Lu, D.~F. Williamson, G.~Jaume, A.~H. Song, B.~Chen, A.~Zhang, D.~Shao, M.~Shaban \emph{et~al.}, ``Towards a general-purpose foundation model for computational pathology,'' \emph{Nature Medicine}, vol.~30, no.~3, pp. 850--862, 2024.

\bibitem{song2023artificial}
A.~H. Song, G.~Jaume, D.~F. Williamson, M.~Y. Lu, A.~Vaidya, T.~R. Miller, and F.~Mahmood, ``Artificial intelligence for digital and computational pathology,'' \emph{Nature Reviews Bioengineering}, vol.~1, no.~12, pp. 930--949, 2023.

\bibitem{lipkova2022artificial}
J.~Lipkova, R.~J. Chen, B.~Chen, M.~Y. Lu, M.~Barbieri, D.~Shao, A.~J. Vaidya, C.~Chen, L.~Zhuang, D.~F. Williamson \emph{et~al.}, ``Artificial intelligence for multimodal data integration in oncology,'' \emph{Cancer cell}, vol.~40, no.~10, pp. 1095--1110, 2022.

\bibitem{jiang2023deep}
H.~Jiang, Y.~Zhou, Y.~Lin, R.~C. Chan, J.~Liu, and H.~Chen, ``Deep learning for computational cytology: A survey,'' \emph{Medical Image Analysis}, vol.~84, p. 102691, 2023.

\bibitem{kather2019deep}
J.~N. Kather, A.~T. Pearson, N.~Halama, D.~J{\"a}ger, J.~Krause, S.~H. Loosen, A.~Marx, P.~Boor, F.~Tacke, U.~P. Neumann \emph{et~al.}, ``Deep learning can predict microsatellite instability directly from histology in gastrointestinal cancer,'' \emph{Nature medicine}, vol.~25, no.~7, pp. 1054--1056, 2019.

\bibitem{chen2022pan}
R.~J. Chen, M.~Y. Lu, D.~F. Williamson, T.~Y. Chen, J.~Lipkova, Z.~Noor, M.~Shaban, M.~Shady, M.~Williams, B.~Joo \emph{et~al.}, ``Pan-cancer integrative histology-genomic analysis via multimodal deep learning,'' \emph{Cancer Cell}, vol.~40, no.~8, pp. 865--878, 2022.

\bibitem{amores2013multiple}
J.~Amores, ``Multiple instance classification: Review, taxonomy and comparative study,'' \emph{Artificial intelligence}, vol. 201, pp. 81--105, 2013.

\bibitem{kanavati2020weakly}
F.~Kanavati, G.~Toyokawa, S.~Momosaki, M.~Rambeau, Y.~Kozuma, F.~Shoji, K.~Yamazaki, S.~Takeo, O.~Iizuka, and M.~Tsuneki, ``Weakly-supervised learning for lung carcinoma classification using deep learning,'' \emph{Scientific reports}, vol.~10, no.~1, p. 9297, 2020.

\bibitem{cao2023e2efp}
L.~Cao, J.~Wang, Y.~Zhang, Z.~Rong, M.~Wang, L.~Wang, J.~Ji, Y.~Qian, L.~Zhang, H.~Wu \emph{et~al.}, ``E2efp-mil: End-to-end and high-generalizability weakly supervised deep convolutional network for lung cancer classification from whole slide image,'' \emph{Medical Image Analysis}, vol.~88, p. 102837, 2023.

\bibitem{li2021dt}
H.~Li, F.~Yang, Y.~Zhao, X.~Xing, J.~Zhang, M.~Gao, J.~Huang, L.~Wang, and J.~Yao, ``Dt-mil: deformable transformer for multi-instance learning on histopathological image,'' in \emph{Medical Image Computing and Computer Assisted Intervention--MICCAI 2021: 24th International Conference, Strasbourg, France, September 27--October 1, 2021, Proceedings, Part VIII 24}.\hskip 1em plus 0.5em minus 0.4em\relax Springer, 2021, pp. 206--216.

\bibitem{lu2021data}
M.~Y. Lu, D.~F. Williamson, T.~Y. Chen, R.~J. Chen, M.~Barbieri, and F.~Mahmood, ``Data-efficient and weakly supervised computational pathology on whole-slide images,'' \emph{Nature biomedical engineering}, vol.~5, no.~6, pp. 555--570, 2021.

\bibitem{li2021dual}
B.~Li, Y.~Li, and K.~W. Eliceiri, ``Dual-stream multiple instance learning network for whole slide image classification with self-supervised contrastive learning,'' in \emph{Proceedings of the IEEE/CVF conference on computer vision and pattern recognition}, 2021, pp. 14\,318--14\,328.

\bibitem{gu2023mamba}
A.~Gu and T.~Dao, ``Mamba: Linear-time sequence modeling with selective state spaces,'' \emph{arXiv preprint arXiv:2312.00752}, 2023.

\bibitem{yang2024mambamil}
S.~Yang, Y.~Wang, and H.~Chen, ``Mambamil: Enhancing long sequence modeling with sequence reordering in computational pathology,'' \emph{arXiv preprint arXiv:2403.06800}, 2024.

\bibitem{fang2024mammil}
Z.~Fang, Y.~Wang, Z.~Wang, J.~Zhang, X.~Ji, and Y.~Zhang, ``Mammil: Multiple instance learning for whole slide images with state space models,'' \emph{arXiv preprint arXiv:2403.05160}, 2024.

\bibitem{dao2024transformers}
T.~Dao and A.~Gu, ``Transformers are ssms: Generalized models and efficient algorithms through structured state space duality,'' \emph{arXiv preprint arXiv:2405.21060}, 2024.

\bibitem{van2021deep}
J.~Van~der Laak, G.~Litjens, and F.~Ciompi, ``Deep learning in histopathology: the path to the clinic,'' \emph{Nature medicine}, vol.~27, no.~5, pp. 775--784, 2021.

\bibitem{shmatko2022artificial}
A.~Shmatko, N.~Ghaffari~Laleh, M.~Gerstung, and J.~N. Kather, ``Artificial intelligence in histopathology: enhancing cancer research and clinical oncology,'' \emph{Nature cancer}, vol.~3, no.~9, pp. 1026--1038, 2022.

\bibitem{lerousseau2020weakly}
M.~Lerousseau, M.~Vakalopoulou, M.~Classe, J.~Adam, E.~Battistella, A.~Carr{\'e}, T.~Estienne, T.~Henry, E.~Deutsch, and N.~Paragios, ``Weakly supervised multiple instance learning histopathological tumor segmentation,'' in \emph{Medical Image Computing and Computer Assisted Intervention--MICCAI 2020: 23rd International Conference, Lima, Peru, October 4--8, 2020, Proceedings, Part V 23}.\hskip 1em plus 0.5em minus 0.4em\relax Springer, 2020, pp. 470--479.

\bibitem{chikontwe2020multiple}
P.~Chikontwe, M.~Kim, S.~J. Nam, H.~Go, and S.~H. Park, ``Multiple instance learning with center embeddings for histopathology classification,'' in \emph{Medical Image Computing and Computer Assisted Intervention--MICCAI 2020: 23rd International Conference, Lima, Peru, October 4--8, 2020, Proceedings, Part V 23}.\hskip 1em plus 0.5em minus 0.4em\relax Springer, 2020, pp. 519--528.

\bibitem{li2023task}
H.~Li, C.~Zhu, Y.~Zhang, Y.~Sun, Z.~Shui, W.~Kuang, S.~Zheng, and L.~Yang, ``Task-specific fine-tuning via variational information bottleneck for weakly-supervised pathology whole slide image classification,'' in \emph{Proceedings of the IEEE/CVF Conference on Computer Vision and Pattern Recognition}, 2023, pp. 7454--7463.

\bibitem{lu2023visual}
M.~Y. Lu, B.~Chen, A.~Zhang, D.~F. Williamson, R.~J. Chen, T.~Ding, L.~P. Le, Y.-S. Chuang, and F.~Mahmood, ``Visual language pretrained multiple instance zero-shot transfer for histopathology images,'' in \emph{Proceedings of the IEEE/CVF conference on computer vision and pattern recognition}, 2023, pp. 19\,764--19\,775.

\bibitem{song2024morphological}
A.~H. Song, R.~J. Chen, T.~Ding, D.~F. Williamson, G.~Jaume, and F.~Mahmood, ``Morphological prototyping for unsupervised slide representation learning in computational pathology,'' in \emph{Proceedings of the IEEE/CVF Conference on Computer Vision and Pattern Recognition}, 2024, pp. 11\,566--11\,578.

\bibitem{shao2021transmil}
Z.~Shao, H.~Bian, Y.~Chen, Y.~Wang, J.~Zhang, X.~Ji \emph{et~al.}, ``Transmil: Transformer based correlated multiple instance learning for whole slide image classification,'' \emph{Advances in neural information processing systems}, vol.~34, pp. 2136--2147, 2021.

\bibitem{chen2021multimodal}
R.~J. Chen, M.~Y. Lu, W.-H. Weng, T.~Y. Chen, D.~F. Williamson, T.~Manz, M.~Shady, and F.~Mahmood, ``Multimodal co-attention transformer for survival prediction in gigapixel whole slide images,'' in \emph{Proceedings of the IEEE/CVF international conference on computer vision}, 2021, pp. 4015--4025.

\bibitem{xiong2021nystromformer}
Y.~Xiong, Z.~Zeng, R.~Chakraborty, M.~Tan, G.~Fung, Y.~Li, and V.~Singh, ``Nystr{\"o}mformer: A nystr{\"o}m-based algorithm for approximating self-attention,'' in \emph{Proceedings of the AAAI Conference on Artificial Intelligence}, vol.~35, no.~16, 2021, pp. 14\,138--14\,148.

\bibitem{vaswani2017attention}
A.~Vaswani, N.~Shazeer, N.~Parmar, J.~Uszkoreit, L.~Jones, A.~N. Gomez, {\L}.~Kaiser, and I.~Polosukhin, ``Attention is all you need,'' \emph{Advances in neural information processing systems}, vol.~30, 2017.

\bibitem{zhu2024vision}
L.~Zhu, B.~Liao, Q.~Zhang, X.~Wang, W.~Liu, and X.~Wang, ``Vision mamba: Efficient visual representation learning with bidirectional state space model,'' \emph{arXiv preprint arXiv:2401.09417}, 2024.

\bibitem{lu2024videomambapro}
H.~Lu, A.~A. Salah, and R.~Poppe, ``Videomambapro: A leap forward for mamba in video understanding,'' \emph{arXiv preprint arXiv:2406.19006}, 2024.

\bibitem{brancati2022bracs}
N.~Brancati, A.~M. Anniciello, P.~Pati, D.~Riccio, G.~Scognamiglio, G.~Jaume, G.~De~Pietro, M.~Di~Bonito, A.~Foncubierta, G.~Botti \emph{et~al.}, ``Bracs: A dataset for breast carcinoma subtyping in h\&e histology images,'' \emph{Database}, vol. 2022, p. baac093, 2022.

\bibitem{ilse2018attention}
M.~Ilse, J.~Tomczak, and M.~Welling, ``Attention-based deep multiple instance learning,'' in \emph{International conference on machine learning}.\hskip 1em plus 0.5em minus 0.4em\relax PMLR, 2018, pp. 2127--2136.

\bibitem{fillioux2023structured}
L.~Fillioux, J.~Boyd, M.~Vakalopoulou, P.-H. Courn{\`e}de, and S.~Christodoulidis, ``Structured state space models for multiple instance learning in digital pathology,'' in \emph{International Conference on Medical Image Computing and Computer-Assisted Intervention}.\hskip 1em plus 0.5em minus 0.4em\relax Springer, 2023, pp. 594--604.

\bibitem{he2016deep}
K.~He, X.~Zhang, S.~Ren, and J.~Sun, ``Deep residual learning for image recognition,'' in \emph{Proceedings of the IEEE conference on computer vision and pattern recognition}, 2016, pp. 770--778.

\bibitem{deng2009imagenet}
J.~Deng, W.~Dong, R.~Socher, L.-J. Li, K.~Li, and L.~Fei-Fei, ``Imagenet: A large-scale hierarchical image database,'' in \emph{2009 IEEE conference on computer vision and pattern recognition}.\hskip 1em plus 0.5em minus 0.4em\relax Ieee, 2009, pp. 248--255.

\end{thebibliography}
\end{document}